\documentclass{article}

\usepackage[square, numbers]{natbib}

\usepackage[preprint]{main_2025}
\usepackage{enumitem}

\usepackage{soul}
\usepackage[utf8]{inputenc} 
\usepackage[T1]{fontenc}    
\usepackage{url}            
\usepackage{booktabs}       
\usepackage{amsfonts}       
\usepackage{nicefrac}       
\usepackage{microtype}      

\usepackage[format=plain,labelformat=simple,labelsep=period,font=small,compatibility=false]{caption}
\usepackage{graphicx}
\usepackage{subfloat}
\usepackage{amsmath}
\usepackage{amssymb}
\usepackage[bb=dsserif]{mathalpha}  

\usepackage{caption}
\usepackage{subcaption}
\usepackage{afterpage}  
\usepackage{multirow}
\usepackage{makecell}
\usepackage{placeins}  
\usepackage{ifthen}  
\usepackage{diagbox}  
\usepackage[hang,flushmargin,bottom]{footmisc}  
\usepackage[table,xcdraw]{xcolor}
\usepackage[export]{adjustbox}
\usepackage[autostyle, english = american]{csquotes}
\MakeOuterQuote{"}  

\definecolor{citecolor}{HTML}{0071bc}
\usepackage[pagebackref,breaklinks,colorlinks,citecolor=citecolor]{hyperref}
\usepackage[capitalize]{cleveref}
\crefname{section}{Sec.}{Secs.}
\Crefname{section}{Section}{Sections}
\Crefname{table}{Table}{Tables}
\crefname{table}{Tab.}{Tabs.}
\crefname{algorithm}{Alg.}{Algs.}

\definecolor{darkred}{HTML}{ea4335}  
\definecolor{green}{HTML}{39b54a}  

\newcommand{\up}[2][]{%
  \ifthenelse { \equal {#1} {} }
  {\color{red}\fontsize{7pt}{1em}\selectfont $\uparrow$#2}
  {\color{#1}\fontsize{7pt}{1em}\selectfont $\uparrow$#2}
}

\newlength\savewidth

\renewcommand{\paragraph}[1]{\vspace{.15mm}\noindent\textbf{#1}}

\newcolumntype{x}[1]{>{\centering\arraybackslash}p{#1pt}}
\newcolumntype{y}[1]{>{\raggedright\arraybackslash}p{#1pt}}
\newcolumntype{z}[1]{>{\raggedleft\arraybackslash}p{#1pt}}

\newcommand{\app}{\raise.17ex\hbox{$\scriptstyle\sim$}}

\definecolor{deemph}{gray}{0.6}

\definecolor{baselinecolor}{gray}{.9}

\title{Holistic Large-Scale Scene Reconstruction \\ via Mixed Gaussian Splatting}

\author{
Chuandong Liu\footnotemark[1]\  , \ 
Huijiao Wang\footnotemark[3]\  , \ 
Lei Yu\footnotemark[2]\ ~\footnotemark[3] , \
Gui-Song Xia\footnotemark[1]\ ~\footnotemark[2] ~\footnotemark[4] ~\footnotemark[5]\ ~\footnotemark[1]\\\
\footnotemark[1] \ School of Computer Science, Wuhan University \\
\footnotemark[2] \ School of Artificial Intelligence, Wuhan University \\
\footnotemark[3] \ School of Electronic Information, Wuhan University \\
\footnotemark[4] \ State Key Lab. of LIESMARS, Wuhan University \\
\footnotemark[5] \ Institute for Math \& AI, Wuhan University \\
}

\begin{document}

\maketitle

\begin{abstract}
   Recent advances in 3D Gaussian Splatting have shown remarkable potential for novel view synthesis. However, most existing large-scale scene reconstruction methods rely on the divide-and-conquer paradigm, which often leads to the loss of global scene information and requires complex parameter tuning due to scene partitioning and local optimization. To address these limitations, we propose MixGS, a novel holistic optimization framework for large-scale 3D scene reconstruction. MixGS models the entire scene holistically by integrating camera pose and Gaussian attributes into a view-aware representation, which is decoded into fine-detailed Gaussians. Furthermore, a novel mixing operation combines decoded and original Gaussians to jointly preserve global coherence and local fidelity. Extensive experiments on large-scale scenes demonstrate that MixGS achieves state-of-the-art rendering quality and competitive speed, while significantly reducing computational requirements, enabling large-scale scene reconstruction training on a single 24GB VRAM GPU. The code will be released at \href{https://github.com/azhuantou/MixGS}{https://github.com/azhuantou/MixGS}.
\end{abstract}

\section{Introduction}
\label{sec:intro}

Accurate 3D scene representation is fundamental for a wide range of applications, including aerial surveying~\cite{bozcan2020air}, autonomous driving~\cite{li2019aads,yang2020surfelgan}, environmental monitoring~\cite{LiuC0XL25,abs-2409-17345}, and AR/VR~\cite{GuJLL0SZB23,JiangYXLFWLLG0J24}, all of which demand high-fidelity visual quality and real-time rendering. In recent years, implicit representations such as neural radiance fields (NeRF)~\cite{nerf,xu2023grid,turki2022mega} have achieved impressive results in novel view synthesis, but often suffer from limited detail reconstruction and slow rendering speed. More recently, explicit representations like 3D Gaussian Splatting (3DGS)~\cite{gs} have demonstrated superior performance in both visual quality and computational efficiency, enabling real-time rendering and showing strong adaptability to dynamic scenes~\cite{DBLP:conf/nips/XuFYX24,DBLP:conf/cvpr/YangGZJ0024} and content generation~\cite{zhou2024diffgs,li2024dreammesh4d}.

\begin{figure*}[t]
    \centering
    \includegraphics[width=\textwidth]{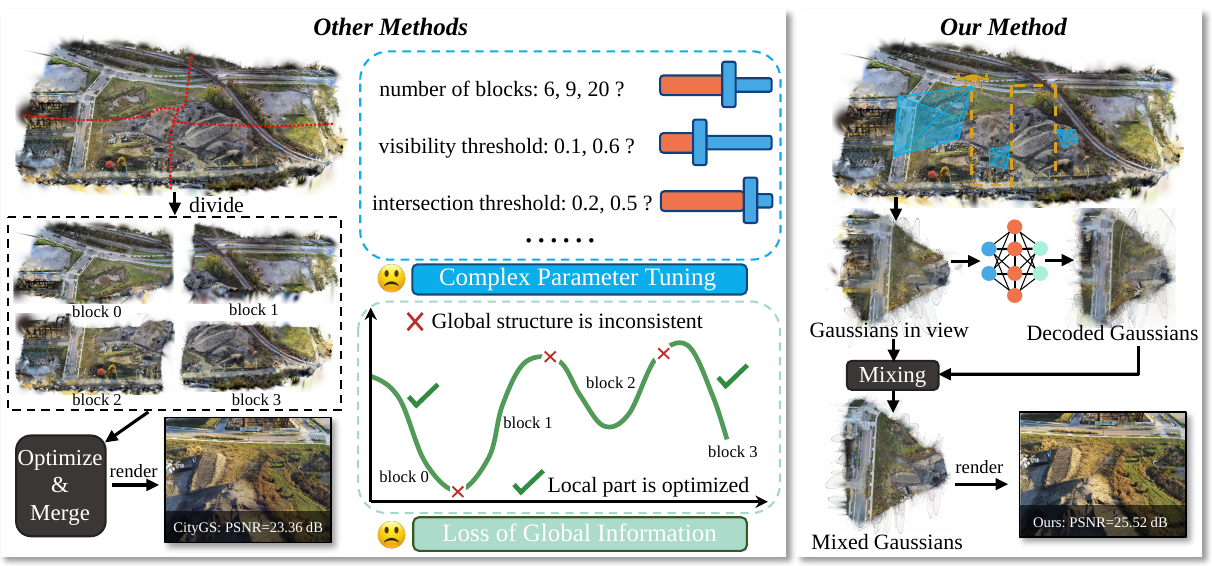}
    \caption{\textbf{Left:} Conventional divide-and-conquer approaches partition the scene into multiple independent blocks, each optimized and rendered separately before merging. This strategy introduces two major limitations: (1) complex parameter tuning, such as selecting the number of blocks and adjusting thresholds (e.g., visibility and intersection~\cite{lin2024vastgaussian,liu2024citygaussian}), which often requires extensive manual intervention and scene-specific reconfiguration; (2) loss of global information, leading to inconsistencies in global structure, illumination, and geometric continuity across block boundaries, as local optimizations do not guarantee global coherence. \textbf{Right:} In contrast, our MixGS framework treats the entire scene and the Gaussian Representation network as a holistic optimization problem. By extracting and encoding visible Gaussians, decoding new Gaussians via implicit feature representations, and mixing them with the originals, MixGS achieves both global consistency and fine-grained detail reconstruction. 
    }
    \label{fig:intro}
    \vspace{-0.3cm}
\end{figure*}

Nevertheless, directly applying naive 3DGS to large-scale scene reconstruction faces significant challenges, such as out-of-memory issues and degraded novel view synthesis quality~\cite{lin2024vastgaussian}. To address these limitations, most existing 3DGS-based methods~\cite{liu2024citygaussian,chen2024dogs} adopt a divide-and-conquer strategy, inspired by its success in previous NeRF-based approaches~\cite{tancik2022block,turki2022mega}. As illustrated in Fig.~\ref{fig:intro} (left), the large scene is partitioned into multiple independent blocks, enabling parallel training and optimization across multiple GPUs. After optimization, these blocks are merged to produce the final reconstruction for novel view rendering. Various improvements, such as progressive partitioning~\cite{lin2024vastgaussian}, recursive scene splitting~\cite{chen2024dogs}, and adaptive data selection~\cite{liu2024citygaussian}, have been proposed to enhance the representation capability of Gaussian primitives within each block. Despite these advances, divide-and-conquer methods still suffer from two major issues.

On the one hand, scene partitioning heavily relies on prior knowledge and manual parameter tuning. The choice of division intervals along the $x$, $y$, and $z$ axes determines the number and size of blocks, which significantly affects the convergence behavior and reconstruction quality. Besides, assigning appropriate training images to each block also requires additional threshold-based adjustments, such as visibility~\cite{lin2024vastgaussian} and intersection~\cite{liu2024citygaussian} thresholds, and these configurations are often scene-specific, necessitating re-tuning for different environments.

On the other hand, independently optimizing each block can lead to locally optimal but globally suboptimal solutions, resulting in the loss of global scene consistency. For instance, lighting conditions that should be uniform across the scene may become inconsistent, and continuous structures like buildings or grasslands can suffer from geometric discontinuities and texture misalignments at block boundaries due to fragmented optimization.

To address the aforementioned limitations of the divide-and-conquer paradigm, \emph{i.e.,} complex parameter tuning and loss of global information, we propose MixGS, a novel 3DGS-based pipeline for large-scale scene reconstruction. Unlike traditional approaches that independently optimize and densify each scene block, as shown in Fig.~\ref{fig:intro} (right), MixGS treats the entire large-scale scene and the Gaussian Representation network as a holistic optimization framework. Specifically, Gaussian primitives within the visible frustum are first extracted based on camera poses and further filtered using opacity information. The positions of these selected Gaussians are encoded via 3D hash encoding to obtain spatial feature representations, while other Gaussian attributes and camera poses are incorporated as auxiliary inputs to enrich the feature space. These features are then decoded to generate refined Gaussians, with a designed offset pool. The decoded Gaussians are mixed with the original filtered Gaussians to form a set of mixed Gaussians for rendering, where the original Gaussians primarily capture coarse scene structure and the decoded Gaussians compensate for missing fine-grained details. This holistic design not only alleviates the high GPU memory requirements by transforming explicit Gaussian densification into implicit high-dimensional feature encoding, enabling training on a single 24GB GPU, but also achieves rendering speeds comparable to divide-and-conquer based methods.

To comprehensively evaluate our approach, we conduct extensive experiments on two prominent benchmarks~\cite{u3d,turki2022mega} for large-scale 3D scene reconstruction, encompassing four challenging scenes. Our results demonstrate that MixGS significantly improves reconstruction performance in these scenarios. To sum up, our key contributions are as follow:
  
\begin{itemize}[nosep,leftmargin=*]
\item We introduce MixGS, a novel holistic optimization method designed to overcome the limitations of existing approaches that rely on divide-and-conquer strategies for large-scale scene reconstruction.

\item We develop a mixed Gaussians rasterization pipeline that simultaneously captures global and local scene information through view-aware Representation modeling for implicit feature learning.

\item Extensive experimentation demonstrate that MixGS achieves state-of-the-art results while maintaining comparable rendering speed on established benchmarks for large-scale scene reconstruction.
\end{itemize}

\section{Related Work}
\label{sec:Related Work}

\textbf{Neural Radiance Fields.}
Neural radiance fields~\cite{nerf} (NeRF) revolutionized 3D scene representation by learning a continuous volumetric function that maps spatial coordinates to color and density values through ray sampling. This breakthrough has inspired numerous follow-up workss~\cite{DBLP:conf/eccv/ChenXGYS22,DBLP:conf/cvpr/Fridovich-KeilY22,DBLP:conf/iccv/YuLT0NK21} that address its computational and representational limitations. Recent advances have focused on improving training efficiency and rendering quality through various scene encoding strategies, including sparse voxel representations~\cite{DBLP:conf/nips/LiuGLCT20}, multi-resolution hash tables~\cite{hash}, and orthogonal axis-plane decompositions~\cite{DBLP:conf/nips/NamRKP23}. These approaches have significantly enhanced NeRF's practical applicability while maintaining its high-quality rendering capabilities.
Beyond per-scene reconstruction, significant progress has been made in generalizable NeRF approaches~\cite{DBLP:conf/iccv/ChenXZZXY021,DBLP:conf/cvpr/JohariLF22,DBLP:conf/cvpr/LiuPLWWTZW22,DBLP:conf/cvpr/SuhailESM22,DBLP:conf/cvpr/WangWGSZBMSF21}, which aim to transfer learned priors across different scenes. These methods often incorporate additional supervision signals, such as depth information~\cite{DBLP:conf/cvpr/DengLZR22,DBLP:conf/cvpr/RoessleBMSN22,DBLP:conf/iccv/WeiLRZL021}, to improve training stability and reconstruction quality. To address the inherent aliasing artifacts in NeRF, researchers have proposed various solutions, including cone tracing approximations with scale-aware positional encodings~\cite{DBLP:conf/iccv/BarronMTHMS21,DBLP:conf/cvpr/BarronMVSH22} and hexagonal sampling strategies~\cite{hash,DBLP:conf/iccv/BarronMVSH23}. For large-scale scene reconstruction, recent works have explored the alignment of multiple NeRF blocks through both traditional optimization techniques~\cite{DBLP:conf/icra/GoliRSGT23} and geometry-aware transformers~\cite{DBLP:conf/iccv/ChenL23} pretrained on 3D datasets.

\textbf{3D Gaussian Splatting.}
3D Gaussian Splatting (3DGS)~\cite{gs} represents a paradigm shift from NeRF's volumetric rendering approach, offering a more efficient explicit representation through differentiable rasterization of 3D Gaussians. While 3DGS achieves real-time rendering performance, it faces challenges in identifying and optimizing Gaussian primitives, particularly in textureless regions. Recent works have addressed this limitation through various innovative approaches. Scaffold-GS~\cite{scaffoldgs} introduces a sparse voxel grid representation to encode Gaussian characteristics, effectively reducing unnecessary densification while maintaining scene fidelity. Further advancements include implicit scene encoding through graph neural networks (SAGS~\cite{sags}) and hierarchical level-of-detail structures (Octree-GS~\cite{octreegs}).
The aliasing issue, common to both NeRF and 3DGS due to fixed-window Gaussian kernels during rasterization, has been addressed in Mip-Splatting~\cite{mipspla} and subsequent works~\cite{DBLP:conf/eccv/LiangZHZFJ24,DBLP:journals/corr/abs-2403-19615,DBLP:conf/cvpr/YanLCL24,DBLP:journals/corr/abs-2411-18625}. Additional research directions include 2D Gaussian~\cite{huang20242d} learning for improved surface fitting~\cite{liu2024citygaussianv2}, dynamic scene modeling through deformation learning~\cite{DBLP:conf/nips/XuFYX24,DBLP:conf/cvpr/YangGZJ0024,wangdc}, and model compression through quantization techniques~\cite{DBLP:conf/nips/FanWWZXW24,DBLP:conf/nips/WangLGYKW24}.

\textbf{Large-Scale Scene Reconstruction.}
Large-scale scene reconstruction has evolved significantly from early Structure-from-Motion (SfM) approaches~\cite{DBLP:journals/tog/SnavelySS06,DBLP:conf/cvpr/SchonbergerF16}, which relied on traditional keypoint extraction methods like SIFT~\cite{DBLP:journals/ijcv/Lowe04}. The availability of camera poses has facilitated scene partitioning, making the divide-and-conquer strategy a mainstream approach for large-scale reconstruction. This paradigm has been successfully adopted in NeRF-based methods~\cite{li2024nerf,rematas2022urban,turki2023suds,xu2023grid,zhang2025efficient}, with notable contributions including Block-NeRF~\cite{tancik2022block}, which incorporates appearance embeddings~\cite{martin2021nerf} and learnable pose refinement for handling environmental variations. Switch-NeRF~\cite{zhenxing2022switch} and Mega-NeRF~\cite{turki2022mega} further enhance this approach through switch transformer-based ray assignment and network sparsity optimization, respectively.

The recent adoption of 3D Gaussian Splatting~\cite{gs} has brought new possibilities to large-scale reconstruction, offering an optimal balance between visual fidelity~\cite{li2024retinags,fan2024momentum,zhang2024geolrm,wu2025blockgaussian} and rendering speed~\cite{chen2024gigags,feng2024flashgs,renscube}. Following the successful divide-and-conquer strategy from NeRF, methods like CityGaussian~\cite{liu2024citygaussian} and VastGaussian~\cite{lin2024vastgaussian} have developed specialized blocking strategies for 3DGS. These approaches ensure optimal training image distribution within blocks, enabling efficient techniques such as Level of Detail (LOD)~\cite{kerbl2024hierarchical,liu2024citygaussianv2,cui2024letsgo}. DOGS~\cite{chen2024dogs} introduces a recursive splitting approach for balanced block sizes, coupled with distributed training for accelerated optimization. While these methods have demonstrated impressive progress in large-scale scene reconstruction, they still face two fundamental challenges: the need for complex parameter tuning and the potential loss of global scene consistency, which our work aims to address.

\section{Approach}
\subsection{Problem Define and Formulation}
\label{sec:3.1}

3D Gaussian Splatting~\cite{gs} explicitly represents a static scene as a set of anisotropic 3D Gaussians $\mathcal{G}=\{ G_i\}_{i=1}^N$, offering both high visual fidelity and efficient training. Each Gaussian kernel is parameterized by its center $\mu_i \in \mathbb{R}^{3}$ and covariance matrix $\Sigma_i \in \mathbb{R}^{3 \times 3}$, where $i$ indexes the $i$-th Gaussian. The 3D Gaussian function at a point $\mathbf{x} \in \mathbb{R}^{3}$ is defined as:
\begin{align} \label{gs}
{G}_{i}(\mathbf{x}) = e^{-\frac{1}{2} (\mathbf{x} - \mu_{i})^{\mathsf{T}} \mathbf{\Sigma}_{i}^{-1} (\mathbf{x} - \mu_{i})},
\end{align}
where the covariance matrix $\Sigma_i$ is further factorized as $\Sigma_i = R_i S_i S_i^{\top} R_i^{\top}$, with $R_i \in \mathbb{R}^{3 \times 3}$ as the rotation matrix and $S_i \in \mathbb{R}^{3 \times 3}$ as the scale matrix.
To render an image from the 3D Gaussian representation, the Gaussians are transformed from world to camera coordinates via a view transformation matrix $W \in \mathbb{R}^{3 \times 3}$ and projected onto the 2D image plane using a Jacobian matrix $J \in \mathbb{R}^{3 \times 3}$~\cite{yifan2019differentiable,zwicker2001surface}. The covariance matrix in the camera plane is computed as $\mathbf{\Sigma}' = \mathbf{J} \mathbf{W} \mathbf{\Sigma} \mathbf{W}^\top \mathbf{J}^\top$.
Given a pixel $\mathbf{p}$ on the image plane, the final color is rendered by compositing the splatted Gaussians from front to back, following the conventional volume rendering equation:
\begin{align} \label{gs2}
\mathbf{C}(\mathbf{p}) = \sum_{i} \mathbf{c}_i \alpha_i \prod_{j=1}^{i-1} (1 - \alpha_j),
\end{align}
where $\mathbf{c}_i$ is the color of the $i$-th Gaussian (computed from spherical harmonic coefficients), and $\alpha_i$ is the opacity associated with $G_{i}$. Each Gaussian $G_i$ is thus parameterized as $(\mu_i, \alpha_i, r_i, s_i, c_i)$, where $r_i$ is the quaternion corresponding to $R_i$, $s_i$ is the scale vector (the diagonal of $S_i$), and $c_i$ is the color.

Based on the above concept, for small-scale scene reconstruction, Gaussians can be directly optimized via the simplified objective function:
\begin{align} \label{gs_my}
\min_{\mathcal{G}} \sum_{i=1}^{M} \mathcal{L}\left(\xi(\mathcal{G}, {\tau_{i}}), I_i^{\mathrm{gt}}\right),
\end{align}
where $\mathcal{G}$ denotes the set of Gaussians that  need to be optimized, $I^{\mathrm{gt}}$ is the set of ground truth images, $\tau_{i}$ represents camera poses, $M$ is the total number of training images, $\xi$ is the rendering process, and $\mathcal{L}$ is the loss function. As defined in~\cite{gs}, $\mathcal{L} = \mathcal{L}_1 + \lambda \mathcal{L}_{\mathrm{SSIM}}$.
However, directly training large-scale scenes with naive 3DGS is limited by memory constraints. 

\textbf{Divide-and-conquer schema:} To address the limitations of naive 3DGS for large-scale scene, most existing methods adopt the divide-and-conquer strategy~\cite{turki2022mega,liu2024citygaussian}. Specifically, the set of Gaussians $\mathcal{G}$ is partitioned into $n$ blocks, \emph{i.e.,} $\mathcal{G} \rightarrow{} \{ \mathcal{G}_{b1}, \dots, \mathcal{G}_{bn} \}$, and each block is assigned a corresponding subset of training images: $I^{\mathrm{gt}} \rightarrow{} \{ I^{\mathrm{gt}}_{b1}, \dots, I^{\mathrm{gt}}_{bn}\}$.
Next, each block is optimized independently:
\begin{align} \label{gs_op}
\mathcal{G}_{bj}^* = \min_{\mathcal{G}_{bj}} \sum_{i=1}^{M_{bj}} \mathcal{L}\left(\xi({\mathcal{G}_{bj}}, {\tau_{i}}), I^{\mathrm{gt}}_{i,bj}\right), \quad \forall bj \in \{b1, \dots, bn\},
\end{align}
where ${M_{bj}}$ denotes the number of training images for block $bj$.
Finally, all optimized blocks are merged to produce the final set of Gaussians for rendering.
While divide-and-conquer methods have achieved success through sophisticated partitioning strategies, they still face two major limitations: complex parameter tuning and loss of global scene information.

\textbf{Our MixGS schema:} In contrast, we propose a holistic optimization framework that jointly models the entire scene, aiming to design an objective function capable of optimizing large-scale scenes in the same manner as small-scale ones.
Specifically, we pretrain coarse Gaussians $\mathcal{G}_c$ from the original large-scale scene and employ a view-aware representations to model the feature attributes under the current viewpoint. Further, the representations with learnable parameter $\Phi$ can be decoded to Gaussian primitives $\mathcal{G}_\Phi$ enriched with scene details, which are then mixed with the coarse Gaussians for final rendering. To sum up, our approach can be formulated as the following objective:
\begin{align} \label{gs_mixgs}
\mathcal{G}_c^*, \Phi^* = \min_{\mathcal{G}_c, \Phi} \sum_{i=1}^M \mathcal{L}\left(\xi({\mathcal{G}_c} \cup \mathcal{G}_\Phi, {\tau_{i}}), I_i^{\mathrm{gt}}\right). 
\end{align}
It is easy to see that we are able to optimize large-scale scenes holistically in the same manner as small-scale scenes.
Besides, we achieve high-quality rendering that captures both global structure and local details, thereby eliminating the need for complex parameter tuning and mitigating the loss of global information, while enabling training on a single 24GB consumer-grade GPU.
The following sections detail the design and holistic optimization of Gaussians $\mathcal{G}_c$ and representations obtained from learnable parameters $\Phi$.

\begin{figure*}[t]
    \centering
    \includegraphics[width=13.8cm]{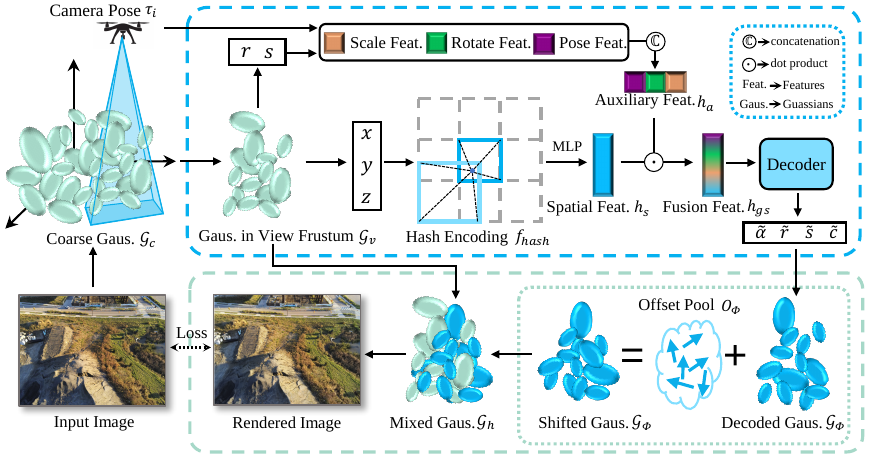}
    \caption{
    Overview of the proposed MixGS method pipeline. We first train the original Gaussians to capture the coarse information of the scene. Then, based on the camera poses, we extract the Gaussians within the view frustum, thereby enabling implicit feature extraction that integrates Gaussian attributes with camera poses. These features are decoded using a tiny multi-head MLP to generate the decoded Gaussians. Next, the positions of the Gaussian primitives are adjusted via an offset pool and combined with the original Gaussians to form the mixed Gaussians. Finally, the mixed Gaussians are splatted through the differentiable rasterization operation~\cite{gs} to render images for supervision.
    }
    \label{fig:pipline}
    \vspace{-0.3cm}
\end{figure*}

\subsection{View-Aware Representation Modeling}
\label{section 3.2}
\textbf{Coarse 3D Gaussians Training.}
To facilitate subsequent feature extraction for different attributes of Gaussians, it is necessary to first capture the coarse geometry and texture information of the large-scale scene. Following the standard 3DGS pipeline~\cite{gs}, we optimize all Gaussians using images and initial point clouds from COLMAP~\cite{DBLP:conf/cvpr/SchonbergerF16}. The resulting set of trained Gaussian primitives is denoted as $\mathcal{G}_{c}=\{ G_i \mid i = 1, \dots, N_c \}$, where $N_c$ represents the total number of Gaussians. This coarse representation provides a strong geometric prior for further attribute learning.

\textbf{Multi-Resolution Hash Encoding.}
Given coarse Gaussians $\mathcal{G}_{c}$, we extract the subset $\mathcal{G}_v=\{ G_i \}_{i=1}^{N_v}$ within the current view frustum, \emph{i.e.,} $\mathcal{G}_{v} \in \mathcal{G}_c$, determined by the camera pose ${\tau_{i}}$ (see in Fig.~\ref{fig:pipline}). The mean positions $\{\mu_i = (x_i, y_i, z_i)\}_{i=1}^{N_v}$ of these Gaussians are used for spatial encoding. We employ multi-resolution hash encoding~\cite{hash} to capture geometric structure at multiple scales:
\begin{align} \label{hash1}
N_l = \left\lfloor N_{\min} \cdot b^l \right\rfloor, \quad 
b = \exp\left( \frac{\ln N_{\max} - \ln N_{\min}}{L-1} \right),
\end{align}
where $N_{\min}$ and $N_{\max}$ represent the coarsest and finest resolutions, respectively, $L$ denotes the largest number of levels, and $l$ is the index of the current level. The selected resolution $N_l$ determines the grid voxel resolution at each level. For a given input $\mathbf{k}$, the corresponding voxel position on the grid can then be computed via rounding down and up $\lfloor \mathbf{k}_l \rfloor = \lfloor \mathbf{k} \cdot N_l \rfloor$, $\lceil \mathbf{k}_l \rceil = \lceil \mathbf{k} \cdot N_l \rceil$. 
To retrieve voxel values, positions are hashed into the corresponding hash table using the following hash function:
\begin{align} \label{hash2}
h_l(\mathbf{k}_l) = \left( \oplus_{i = 1}^{d} k_i \pi_i \right) \mod \quad T,
\end{align}
where $\oplus$ denotes the bit-wise XOR operation, $\{\pi_i\}$ are distinct large prime numbers assigned to each dimension, $d$ represents the input dimensionality, and $T$ indicates the hash table size. The encoded features are then obtained through trilinear interpolation over the corresponding grid voxel values.

Generally, we denote the above hash encoder as $f_{hash}$, the final spatial features $h_s$ are obtained via:
\begin{align} \label{hash3}
h_s = f_{\theta_1} \circ f_{hash}(x_i, y_i, z_i),
\end{align}
where $f_{\theta_1}$ is a tiny MLP with parameters $\theta_1$.
We adopt multi-resolution hash encoding for its hierarchical structure, which enables progressive learning from coarse to fine details for performance gains, and for its efficient implementation~\cite{tiny-cuda-nn}, which ensures fast training and inference.

\textbf{Auxiliary Gaussian Enhancement.}
To further enhance representation capacity, we independently encode the rotation quaternions $r$, scale vectors $s$ of $\mathcal{G}_v$, and the camera pose ${\tau_{i}}$, capturing local geometric variations and viewpoint-dependent effects. These auxiliary features are concatenated to form $h_{a}$ and transformed into attention scores, which modulate the spatial features:
\begin{align} \label{atten}
h_{gs} = (2 \cdot \sigma(f_{\theta_2} (h_a)) - 1) \odot h_s, 
\end{align}
where $h_{gs}$ is the fused feature, $f_{\theta_2}$ is a lightweight MLP, $\odot$ denotes element-wise multiplication, and $\sigma$ is the Sigmoid function.

\subsection{Mixed Gaussians Rasterization}
In this section, the primary objective is to decode the high-dimensional feature $h_{gs}$, which models the view-aware Gaussian attributes, thereby generating corresponding Gaussian attributes $\{\mu, \alpha, r, s, c\}$ as described in Sec. \ref{sec:3.1}. Note that we call the generated Gaussians as decoded Gaussians $\mathcal{G}_\Phi$. In detail, we use a multi-head MLP $f_{\theta_3}$ to decode the features $h_{gs}$ and predict the Gaussian opacity $\tilde{\alpha} \in \mathbb{R}$, color $\tilde{c} \in \mathbb{R}^{3}$, and covariance matrix $\Sigma$, which is reparameterization by a rotation quaternion $\tilde{r} \in \mathbb{R}^{4}$ and a scaling matrix $\tilde{s} \in \mathbb{R}^{3}$. This process can be formulated as:
\begin{align} \label{maskgs1}
\{(\tilde{\alpha}_i, \tilde{r}_i, \tilde{s}_i, \tilde{c}_i) \}_{i=1}^{N_\Phi} = f_{\theta_3}(h_{gs}).
\end{align}
Note that the number of decoded Gaussians are the same with Gaussians in the view frustum, \emph{i.e.,} $N_\Phi=N_v$.
Before being used to formulate Gaussian representations, the rotation quaternion should be normalized and the opacity should be activated by Sigmoid function, to satisfy their range and feature.
Moreover, we design a offset pool $O_c = \{o_1, o_2, \dots, o_{N_c}\}$ to save the shifted value for each decoded Gaussian primitives. Further, the position of decoded Gaussians are calculated as:
\begin{align} \label{maskgs2}
\{\tilde{\mu}_1, \dots, \tilde{\mu}_{N_\Phi} \} = \{\mu_1, \dots, \mu_{N_v} \} + \{o_1, \dots, o_{N_\Phi} \}, 
\end{align}
where $ O_\Phi \in O_c $ is the offset pool in view, and $\mu_{N_v}$ is the position attribute of Gaussians $\mathcal{G}_v$. 
Compared to directly predicting absolute positions, when the coarse Gaussian position is already close to the target structure, fine-tuning through local offset prediction can more effectively capture small-scale geometric variations. By predicting residual deviations rather than complete positional information, the model converges more easily and achieves greater precision in detail refinement.

With the aforementioned decoded Gaussians $\mathcal{G}_\Phi$, a novel view can be rendered using the rasterization splatting pipeline~\cite{gs}. However, the obtained view easily focus on view-aware local details, thereby lacking stable global structure. Hence, we propose to combine Gaussians $\mathcal{G}_\Phi$ and $\mathcal{G}_v$, generating mixed Gaussians $\mathcal{G}_h$ for further rendering. This process can be formulated as:
\begin{align} \label{maskgs3}
\mathcal{G}_{h} = \mathcal{G}_{v} \cup \mathcal{G}_{\Phi} = \{ G_{1}, \ldots, G_{N_v}, G_{N_v+1}, \ldots, G_{N_\Phi+N_v} \}.
\end{align}
Therefore, we can obtain final rendered view via the mixed Gaussians.
In this way, Gaussian densification is implicitly achieved, which facilitates the reconstruction of fine details and complex lighting in large-scale scenes. On the other hand, it significantly reduces the memory requirements during training, enabling efficient training even on a consumer GPU with 24GB of VRAM.

\subsection{Training with Multi-Stage Optimization}
To ensure effective optimization during training, we adopt a three-stage optimization strategy. This approach progressively refines the model from both global and local perspectives, enabling a better balance between global structure and local details in large-scale scene reconstruction.

\textbf{(1) Coarse Stage.} Coarse Gaussians $\mathcal{G}_c$ are trained to rapidly capture the global geometric structure, providing a stable prior and preventing premature overfitting to local details.

\textbf{(2) Detail Stage.} With coarse Gaussian parameters fixed, we optimize other parameters $\Phi$, such as the hash encoder, MLP with learnable parameters, and offset pool, \emph{i.e.,} $\Phi = (f_{\theta_1},f_{\theta_2},f_{\theta_3},f_{hash},O_c)$, allowing the model to fit local details and high-frequency geometry based on the global structure.

\textbf{(3) Joint Stage.} After capturing both global and local information, we jointly fine-tune all parameters ($\mathcal{G}_c$ and $\Phi$) using the mixed Gaussians. This stage integrates global and local cues, eliminates boundary artifacts, and leverages priors from earlier stages, further enhancing overall quality.

\section{Experiments}

\subsection{Experimental Settings}
\textbf{Datasets and Evaluation Metrics.}
Following state-of-the-art methods~\cite{chen2024dogs,liu2024citygaussian}, we conduct experiments on large-scale scenes across two real-world urban scene datasets: UrbanScene3D~\cite{u3d} dataset with Residence and Sci-Art, and Mill19~\cite{turki2022mega} dataset with Building and Rubble. Both of datasets consist of thousands of high-resolution images captured by drones. For fair comparison~\cite{chen2024dogs,turki2022mega,lin2024vastgaussian}, we adopt the same data splits as previous works to construct the training and testing sets. Specifically, the Building, Rubble, Residence, and Sci-Art scenes contain 1920, 1657, 2582, and 3620 training images, respectively, and 20, 21, 21, and 21 testing images.
For the evaluation of novel view synthesis, we report PSNR, SSIM~\cite{ssim}, and LPIPS~\cite{lpips} metrics. Additionally, we measure Frame Per Second (FPS) to assess model effectiveness.

\textbf{Compared Methods.}
We compare MixGS against seven state-of-the-art methods, categorized into NeRF-based methods: Switch-NeRF~\cite{zhenxing2022switch}, Mega-NeRF~\cite{turki2022mega}, GP-NeRF~\cite{xu2023grid} and 3DGS-based methods: VastGaussian~\cite{lin2024vastgaussian}, CityGaussian~\cite{liu2024citygaussian}, Hierarchy-GS~\cite{kerbl2024hierarchical}, DOGS~\cite{chen2024dogs}. To ensure a fair comparison, we downsample all images by 4 times. Note that DOGS~\cite{chen2024dogs} differs slightly from other methods, which downsamples images by a factor of 6 times. Therefore, we also conduct a experiments with 6 times downsampling for direct comparison, as reported in the Appendix.

\textbf{Implementation Details.}
All experiments are conducted on an NVIDIA RTX 3090 GPU with 24GB VRAM using PyTorch. 
Following previous methods~\cite{liu2024citygaussian,chen2024dogs}, we use the official camera poses provided by Mega-NeRF~\cite{turki2022mega} and initialize 3D Gaussians with COLMAP~\cite{DBLP:conf/cvpr/SchonbergerF16}.
The three training stages are run for 30,000, 40,000, and 260,000 iterations, respectively.
To fit the original 3DGS~\cite{gs} for large-scale scene reconstruction, we train for 60,000 iterations, applying densification every 200 steps until 30,000 iterations.
VastGaussian~\cite{lin2024vastgaussian} applies color correction to rendered images before metric evaluation, which can significantly improve results. For fair comparison, we report the results of VastGaussian using a reproduced version~\cite{chen2024dogs} that excludes both color correction and decoupled appearance encoding.

\subsection{Main Results}
\textbf{Quantitative Results.}
Table~\ref{tab:main_result} reports the mean PSNR, SSIM, and LPIPS metrics across the four large-scale scenes. MixGS achieves the highest SSIM across all scenes, highlighting its superior perceptual quality. In terms of PSNR and LPIPS, our method consistently delivers either the best or highly competitive results. Notably, on the challenging Residence dataset, MixGS achieves a PSNR of 23.39, enabling 3DGS-based approaches to outperform NeRF-based methods. It is worth mentioning that superior PSNR values are obtained by NeRF-based methods on the Sci-Art dataset, which exhibits inherent blur due to capture conditions. This can be attributed to the tendency of NeRF-based reconstructions to produce smoother outputs that align with the dataset’s characteristics. In contrast, 3DGS-based methods, including our MixGS, maintain better structural and perceptual fidelity across all scenes, as reflected in the SSIM and LPIPS scores.

\begin{figure}[t]
    \centering
    \includegraphics[width=\columnwidth]{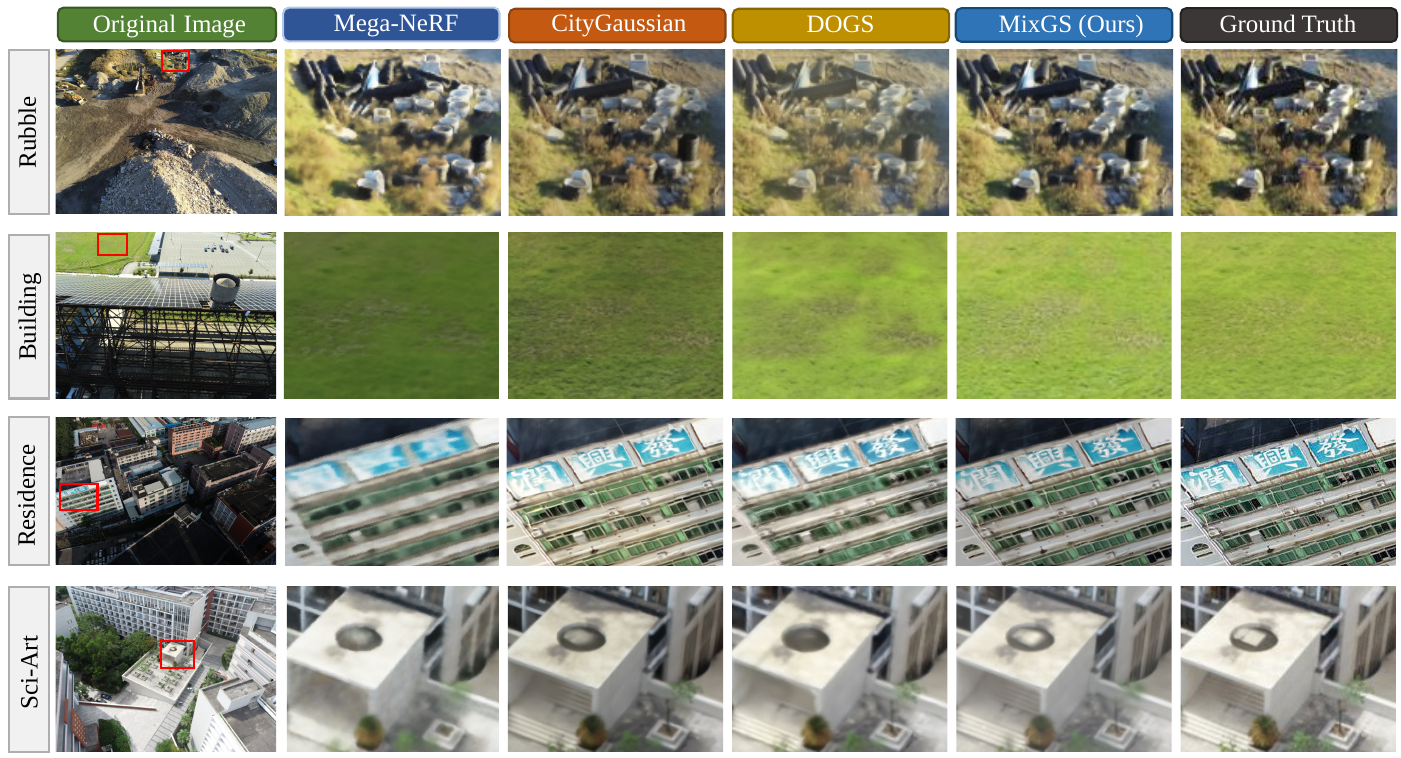}
    \caption{Qualitative comparison of rendering on the Mill19~\cite{turki2022mega} and UrbanScene3D~\cite{u3d} datasets. We demonstrate the zoomed-in detail of the selected areas in red box. }
    \label{fig:compare}
    \vspace{-3mm}
\end{figure}

\textbf{Qualitative Results.}
Fig. \ref{fig:compare} presents the qualitative comparisons of the novel view synthesis results. For MegaNeRF\footnote{https://github.com/cmusatyalab/mega-nerf}, CityGaussian\footnote{https://github.com/Linketic/CityGaussian}, and DOGS\footnote{https://github.com/AIBluefisher/DOGS}, all visualizations are generated using official pretrained models or results. It can be observed that MixGS outperforms existing methods by producing sharper textures, more accurate illumination, and improved geometric consistency. Specifically, MegaNeRF lacks fine details and exhibit blurry and erroneous structures in image rendering. Additionally, MixGS demonstrates significantly better visual quality under challenging lighting conditions, such as the example on the $2^{nd}$ row in Fig. \ref{fig:compare}. Additional visualization results are provided in the Appendix.

\textbf{Rendering Speed.}
To further assess the efficiency of our method, we compare the rendering speed (FPS) with other approaches. As shown in Table.~\ref{tab:fps}, traditional NeRF-based methods, such as Mega-NeRF and Switch-NeRF, exhibit significantly slower rendering speed ($<$0.1 FPS). In contrast, 3DGS-based methods achieve much higher rendering speeds. Benefit from the acceleration provided by the CUDA/C++ implementation and lightweight network design, MixGS achieve real-time speed ($\geq$ 30 FPS) on all large-scale scenes, even when running on a single RTX 3090 GPU, and outperforms or matches methods with more powerful A100 GPUs.
To ensure accurate and fair measurements of rendering times per frame, we explicitly synchronize all CUDA streams prior to timing. 
For more analysis about training time and consumption across methods, please refer to the Appendix.

\begin{table}
\centering
\caption{Quantitative results of our method compared to previous work on Mill19 dataset~\cite{turki2022mega} and UrbanScene3D dataset~\cite{u3d}. We report metrics for PSNR$\uparrow$, SSIM$\uparrow$, and LPIPS$\downarrow$ on test views. The best, second best, and third best results are highlighted in \textcolor{red}{red}, \textcolor{orange}{orange}, and \textcolor{yellow}{yellow} respectively. $\dagger$ represents without applying the decoupled appearance encoding.
}
\label{tab:main_result}
\resizebox{\columnwidth}{!}{
\begin{tabular}{c|ccc|ccc|ccc|ccc}
\midrule
Scenes & \multicolumn{3}{c|}{Building} & \multicolumn{3}{c|}{Rubble} & \multicolumn{3}{c|}{Residence} & \multicolumn{3}{c}{Sci-Art} \\ \midrule
Metrics & PSNR\ $\uparrow$ & SSIM\ $\uparrow$ & LPIPS\ $\downarrow$ & PSNR\ $\uparrow$ & SSIM\ $\uparrow$ & LPIPS\ $\downarrow$ & PSNR\ $\uparrow$ & SSIM\ $\uparrow$ & LPIPS\ $\downarrow$ & PSNR\ $\uparrow$ & SSIM\ $\uparrow$ & LPIPS\ $\downarrow$ \\ \midrule
Mega-NeRF~\cite{turki2022mega} & 20.92 & 0.547 & 0.454 & 24.06 & 0.553 & 0.508 & 22.08 & 0.628 & 0.401 & \cellcolor{orange}25.60 & 0.770 & 0.312 \\
Switch-NeRF~\cite{zhenxing2022switch} & 21.54 & 0.579 & 0.397 & 24.31 & 0.562 & 0.478 & \cellcolor{orange}22.57 & 0.654 & 0.352 & \cellcolor{red}26.51 & 0.795 & 0.271 \\ 
$\text{GP-NeRF}$~\cite{xu2023grid} & 21.03 & 0.566 & 0.486 & 24.06 & 0.565 & 0.496 & \cellcolor{yellow}22.31 & 0.661 & 0.448 & \cellcolor{yellow}25.37 & 0.783 & 0.373 \\ \midrule
$\text{3DGS}$~\cite{gs} & 20.46 & 0.720 & 0.305 & 25.47 & \cellcolor{yellow}0.777 & 0.277 & 21.44 & \cellcolor{yellow}0.791 & 0.236 & 21.05 & \cellcolor{yellow}0.830 & 0.242 \\
$\text{VastGaussian}^{\dagger}$~\cite{lin2024vastgaussian} & \cellcolor{yellow}21.80 & 0.728 & \cellcolor{orange}0.225 & \cellcolor{yellow}25.20 & 0.742 & \cellcolor{yellow}0.264 & 21.01 & 0.699 & 0.261 & 22.64 & 0.761 & 0.261 \\
Hierarchy-GS~\cite{kerbl2024hierarchical} & 21.52 & 0.723 & 0.297 & 24.64 & 0.755 & 0.284 & -- & -- & -- & -- & -- & -- \\
DOGS~\cite{chen2024dogs} & \cellcolor{orange}22.73 & \cellcolor{yellow}0.759 & \cellcolor{red}0.204 & \cellcolor{orange}25.78 & 0.765 & \cellcolor{orange}0.257 & 21.94 & 0.740 & \cellcolor{yellow}0.244 & 24.42 & 0.804 & \cellcolor{red}0.219 \\
CityGaussian~\cite{liu2024citygaussian} & 21.67 & \cellcolor{orange}0.764 & 0.262 & 24.90 & \cellcolor{orange}0.785 & \cellcolor{red}0.256 & 21.90 & \cellcolor{orange}0.805 & \cellcolor{red}0.217 & 21.34 & \cellcolor{orange}0.833 & \cellcolor{yellow}0.232 \\ \midrule
MixGS (Ours) & \cellcolor{red}23.03 & \cellcolor{red}0.771 & \cellcolor{yellow}0.261 & \cellcolor{red}26.66 & \cellcolor{red}0.792 & 0.267 & \cellcolor{red}23.39 & \cellcolor{red}0.815 & \cellcolor{orange}0.219 & 24.20 & \cellcolor{red}0.856 & \cellcolor{orange}0.220 \\ \midrule
\end{tabular}}
\vspace{-3mm}
\end{table}

\subsection{Ablation Study}

Table~\ref{tab:ablation} presents the results of ablation experiments designed to assess the individual contributions of key components in the proposed MixGS framework. We remove or modify each module and evaluate its impact on reconstruction performance, using the Rubble scene as a example.

First, we examine the effect of multi-resolution hash encoding (\emph{w/o} HE). Removing this module results in a substantial performance drop, with PSNR decreasing by approximately 1.96dB compared to the full model. This clearly demonstrates the critical role of hash encoding in representing the 3D positional attributes of Gaussians.
Second, we assess the impact of auxiliary Gaussian enhancement (\emph{w/o} AE). Excluding this component leads to a noticeable decline in performance, primarily due to the absence of camera pose and other Gaussian attribute information. This result highlights the necessity of incorporating auxiliary features to effectively capture the complex variations present in large-scale scenes.
We further validate the effectiveness of the multi-stage training strategy. The $3^{rd}$ and $4^{th}$ rows of Table~\ref{tab:ablation} correspond to the ablation of the first-stage coarse Gaussian training (\emph{w/o} CT) and the second-stage implicit feature training (\emph{w/o} IT), respectively. Omitting either stage prevents the optimization process from converging to an optimal solution, indicating that both global structure initialization and local detail refinement are essential for high-quality reconstruction.
Next, we investigate the role of the offset pool (\emph{w/o} OP), where positional offsets are directly regressed by the decoder rather than learned through a dedicated pool. As shown in the $5^{th}$ row, this leads to a slight decrease in reconstruction performance, as direct regression introduces greater uncertainty and makes it more challenging to capture fine geometric details.

\begin{figure}[t]
    \centering
    \includegraphics[width=\columnwidth]{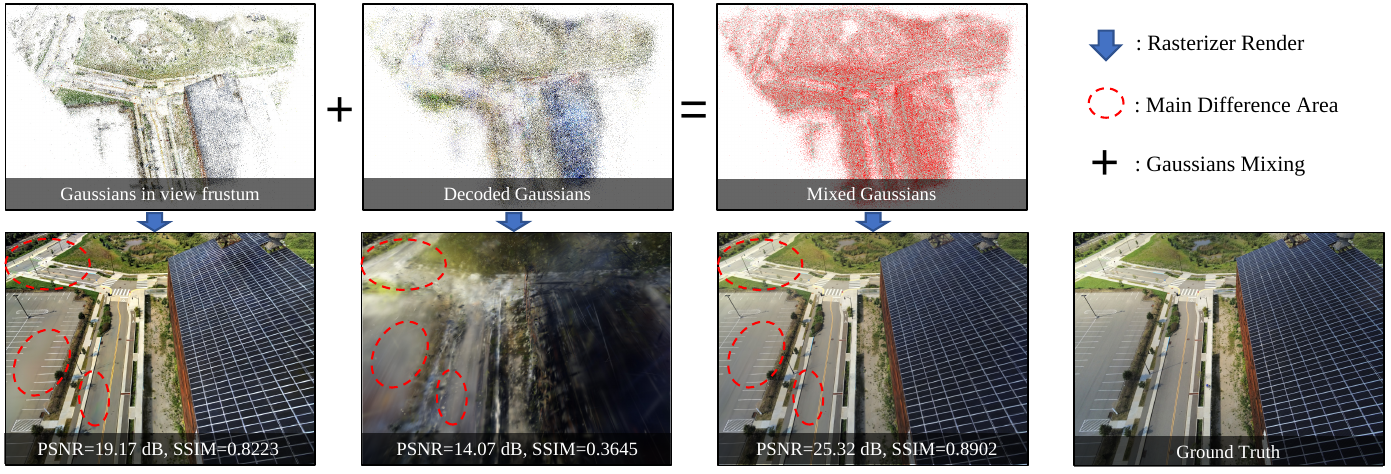}
    \caption{Visualizations of the Gaussians in the view frustum, decoded Gaussians, and mixed Gaussians, along with their corresponding rendered images. It can be observed that rendering with mixed Gaussians leads to better reconstruction of lighting consistency and fine-grained local details.}
    \label{fig:abl}
    \vspace{-4mm}
\end{figure}

Additionally, we conduct an ablation on mixed Gaussian rendering (\emph{w/o} GV), where only the decoded Gaussians are used for rendering, without combining them with the original Gaussians in the view frustum. This configuration results in a significant degradation of reconstruction quality, as relying solely on implicit features makes it difficult to maintain the global structure of large-scale scenes.
To further illustrate the benefits of Gaussian mixing, Fig.~\ref{fig:abl} visualizes the Gaussians in the view frustum, the decoded Gaussians, and the mixed Gaussians, along with their corresponding rendered images. It is evident that rendering with mixed Gaussians leads to superior reconstruction of lighting consistency and fine-grained local details.

\begin{table}[t]
 \parbox{.65\linewidth}{
 \centering
  \caption{Rendering speed comparison on the UrbanScene3D~\cite{u3d} and Mill19~\cite{turki2022mega} datasets. We report the FPS and used GPU type. 
  }
 \label{tab:fps}
 \resizebox{0.65\textwidth}{!}{
\begin{tabular}{c|c|cccc}
\toprule
Methods & GPU Type & Building & Rubble & Residence & Sci-Art \\ \midrule
Mega-NeRF~\cite{turki2022mega} & A100 40G & <0.1 & <0.1 & <0.1 & <0.1 \\
Switch-NeRF~\cite{zhenxing2022switch} & A100 40G & <0.1 & <0.1 & <0.1 & <0.1 \\
GP-NeRF~\cite{xu2023grid} & A100 40G & 0.42 & 0.40 & 0.31 & 0.34 \\ \midrule
3DGS~\cite{gs} & A100 40G & 45.0 & 47.8 & 62.1 & 72.2 \\
CityGaussian (w/ LoD)~\cite{liu2024citygaussian} & A100 40G & 37.4 & 52.6 & 41.6 & 64.6 \\
CityGaussian (w/o LoD)~\cite{liu2024citygaussian} & A100 40G & 24.3 & 43.9 & 32.7 & 56.1 \\
MixGS (Ours) & 3090 24G & 33.5 & 42.5 & 36.6 & 53.9 \\ \bottomrule
\end{tabular}
 }
 }
 \hfill
 \parbox{.34\linewidth}{
 \centering
  \caption{Quantitative ablation results on the \emph{Rubble} dataset. }
 \label{tab:ablation}
 \resizebox{0.34\textwidth}{!}{
\begin{tabular}{c|ccc}
\toprule
Model & PSNR$\uparrow$ & SSIM$\uparrow$ & LPIPS$\downarrow$ \\ \midrule
w/o HE & 24.70 & 0.728 & 0.324 \\
w/o AE & 26.24 & 0.784 & 0.277 \\
w/o CT & 25.86 & 0.763 & 0.307 \\
w/o IT & 25.99 & 0.759 & 0.311 \\
w/o OP & 26.35 & 0.779 & 0.275 \\
w/o GV & 23.65 & 0.704 & 0.360 \\
full model & 26.66 & 0.792 & 0.267 \\ \bottomrule
\end{tabular}
 }
 }
 \vspace{-3mm}
\end{table}

\section{Conclusion}

In this paper, we propose MixGS, a novel framework for large-scale novel view synthesis. In contrast to mainstream approaches that adopt the divide-and-conquer paradigm, MixGS treats the entire scene as a unified whole and performs holistic optimization. Our method introduces a view-aware representation modeling strategy that integrates camera pose and Gaussian attributes to extract informative local features. By leveraging a multi-head decoder and an offset pool, MixGS generates decoded Gaussians that effectively capture fine-grained scene details. The final image rendering is achieved by mixing the Gaussians within the view frustum with the decoded Gaussians, resulting in high-quality scene reconstruction. Extensive experiments on multiple challenging datasets demonstrate that MixGS achieves state-of-the-art performance in view synthesis quality, while maintaining efficiency on a single GPU. Importantly, MixGS provides a fundamentally new perspective for large-scale scene reconstruction, opening up promising directions for future research.

{\small
\bibliographystyle{ieee_fullname}
\bibliography{ref}
}

\newpage
\appendix

\section{Analysis of $6\times$ Downsampling}

\begin{table}[h]
\centering
\caption{Quantitative results of our method compared to previous work on Mill19 dataset~\cite{turki2022mega} and UrbanScene3D dataset~\cite{u3d}. We report metrics for PSNR$\uparrow$, SSIM$\uparrow$, and LPIPS$\downarrow$ on test views. The \emph{best} and \underline{second best} are highlighted.
}
\label{tab:dogs6x}
\resizebox{\columnwidth}{!}{
\begin{tabular}{c|ccc|ccc|ccc|ccc}
\toprule
Scenes & \multicolumn{3}{c|}{Building} & \multicolumn{3}{c|}{Rubble} & \multicolumn{3}{c|}{Residence} & \multicolumn{3}{c}{Sci-Art} \\ \midrule
Metrics & PSNR\ $\uparrow$ & SSIM\ $\uparrow$ & LPIPS\ $\downarrow$ & PSNR\ $\uparrow$ & SSIM\ $\uparrow$ & LPIPS\ $\downarrow$ & PSNR\ $\uparrow$ & SSIM\ $\uparrow$ & LPIPS\ $\downarrow$ & PSNR\ $\uparrow$ & SSIM\ $\uparrow$ & LPIPS\ $\downarrow$ \\ \midrule
DOGS~\cite{chen2024dogs} ($6\times$) & 22.73 & 0.759 & 0.204 & 25.78 & 0.765 & 0.257 & 21.94 & 0.740 & 0.244 & \textbf{24.42} & 0.804 & \underline{0.219} \\
MixGS (Ours) ($4\times$) & \underline{23.03} & \underline{0.771} & \underline{0.261} & \underline{26.66} & \underline{0.792} & \underline{0.267} & \underline{23.39} & \underline{0.815} & \underline{0.219} & 24.20 & \underline{0.856} & 0.220 \\
MixGS (Ours) ($6\times$) & \textbf{23.82} & \textbf{0.8142} & \textbf{0.197} & \textbf{27.95} & \textbf{0846} & \textbf{0.189} & \textbf{23.96} & \textbf{0.859} & \textbf{0.159} & \underline{24.36} & \textbf{0.878} & \textbf{0.1648} \\ \bottomrule
\end{tabular}}
\end{table}

Consistent with most previous works~\cite{turki2022mega,liu2024citygaussian,lin2024vastgaussian}, we primarily downsample all images by a factor of 4, while DOGS~\cite{chen2024dogs} employs a higher downsampling rate of 6 or more. We observe that increasing the downsampling factor to 6 leads to significantly improved results compared to a factor of 4, as the reconstruction task becomes less challenging due to the reduction of high-frequency information. As shown in Table~\ref{tab:dogs6x}, our method achieves substantially better performance when images are downsampled by a factor of 6.

\section{Computational Overhead}

\begin{table}[h]
\centering
\caption{Quantitative results of novel view synthesis on the Mill19 and UrbanScene3D datasets. We report the training time (hh:mm) and the allocated memory (GB). $\dagger$ indicates results obtained without applying decoupled appearance encoding.
}
\label{tab:overhead}
\resizebox{\columnwidth}{!}{
\begin{tabular}{c|c|cc|cc|cc|cc}
\toprule
\multirow{2}{*}{Methods} & \multirow{2}{*}{Used GPU} & \multicolumn{2}{c|}{Building} & \multicolumn{2}{c|}{Rubble} & \multicolumn{2}{c|}{Residence} & \multicolumn{2}{c}{Sci-Art} \\ \cline{3-10} 
 &  & Opt. Time & Mem. & Opt. Time & Mem. & Opt. Time & Mem. & Opt. Time & Mem. \\ \midrule
Mega-NeRF~\cite{turki2022mega} & $5\times$ RTX6000 48GB & 19:49 & 5.84 & 30:48 & 5.88 & 27:20 & 5.99 & 27:39 & 5.97 \\
Switch-NeRF~\cite{zhenxing2022switch} & $5\times$ RTX6000 48GB & 24:46 & 5.84 & 38:30 & 5.87 & 35:11 & 5.94 & 34:34 & 5.92 \\ \midrule
3DGS~\cite{gs} & $1\times$ RTX3090 24GB & 02:39 & 4.62 & 02:24 & 2.18 & 02:49 & 3.23 & 02:03 & 1.61 \\
VastGaussian$\dagger$~\cite{lin2024vastgaussian} & $5\times$ RTX6000 48GB & 03:26 & 3.07 & 02:30 & 2.74 & 03:12 & 3.67 & 02:33 & 3.54 \\
DOGS~\cite{chen2024dogs} & $5\times$ RTX6000 48GB & 03:51 & 3.39 & 02:25 & 2.54 & 04:33 & 6.11 & 04:23 & 3.53 \\
MixGS (Ours) & $1\times$ RTX3090 24GB & 14:39 & 3.25 & 12:07 & 2.76 & 17:09 & 2.83 & 14:48 & 2.24 \\ \bottomrule
\end{tabular}}
\end{table}

To analysis the trade off between different methods, we compare the computational overhead on both the Mill19 and UrbanScene3D datasets. In Table~\ref{tab:overhead}, we report the optimization time, GPU memory consumption during evaluation, and the type of GPU used. For Mega-NeRF, Switch-NeRF, VastGaussian, and DOGS, we directly adopt the corresponding data reported in the original DOGS paper, where training was conducted on five RTX 6000 GPUs with 48GB memory each. For 3DGS and our MixGS, we conduct experiments on a single RTX 3090 GPU paired with an Intel(R) Xeon(R) Gold 6154 CPU @ 3.00GHz, and report the corresponding metrics. The software environment includes PyTorch version 2.0.1 and CUDA version 11.7. 
Traditional NeRF-based methods, Mega-NeRF and Switch-NeRF, exhibit significantly longer optimization times and require substantial computational resources. In contrast, Gaussian-based approaches demonstrate considerably lower optimization times and more efficient memory usage. Compared with other methods, MixGS delivers more stable memory consumption across scenes, with memory usage tightly bounded.

\textbf{Limitations and future work:} Currently, due to the single-GPU setup, our method requires more training time compared to multi-GPU approaches. As a future direction, we plan to adopt the open-source distributed training and accelerate training framework\footnote{https://github.com/nerfstudio-project/gsplat} to reduce the overall time consumption.

\section{More Qualitative Results}
\begin{figure}[t]
    \centering
    \includegraphics[width=\columnwidth]{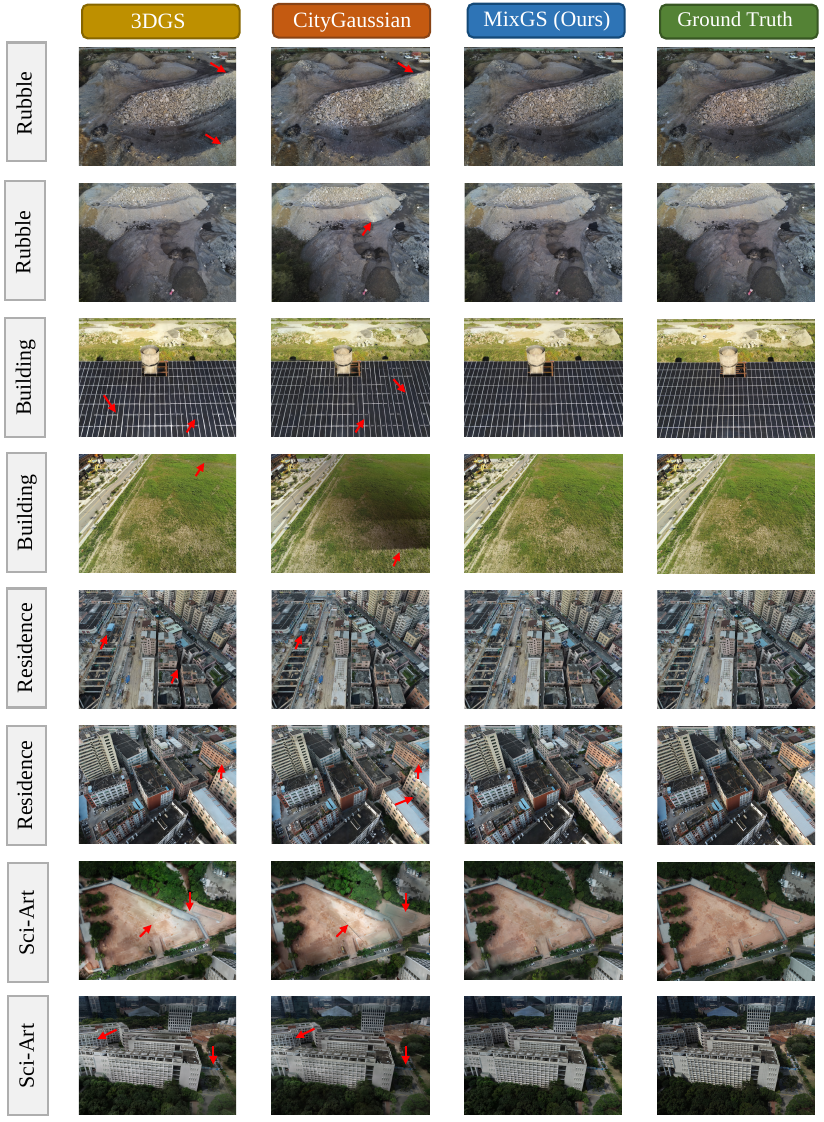}
    \caption{Qualitative results of ours and other methods in image rendering on Mill-19~\cite{turki2022mega} and Urbanscene3D~\cite{u3d} datasets.}
    \label{fig:app_vis}
    \vspace{-3mm}
\end{figure}

We provide additional visual comparisons for the Rubble~\cite{turki2022mega}, Building~\cite{turki2022mega}, Residence~\cite{u3d}, and Sci-Art~\cite{u3d} scenes. Our method consistently reconstructs finer details across all these scenes. Notably, MixGS demonstrates superior capability in reconstructing luminance, as exemplified by the Rubble and Building results shown in Fig.~\ref{fig:app_vis}. For better visualization, we use red arrow to indicate the differences. This highlights the effectiveness of our holistic optimization framework, which prevents local optimum and loss of global information.

\begin{figure}[t]
    \centering
    \includegraphics[width=\columnwidth]{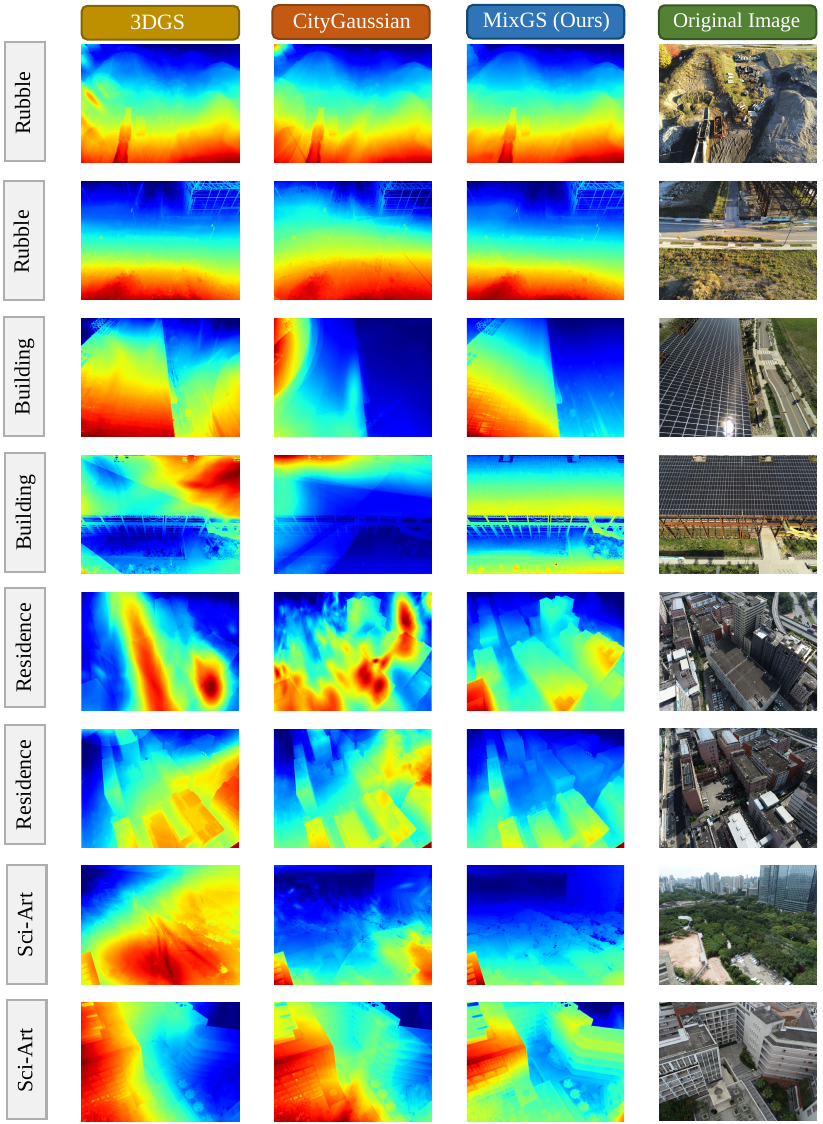}
    \caption{Qualitative results of ours and other methods in depth rendering on Mill-19~\cite{turki2022mega} and Urbanscene3D~\cite{u3d} datasets.}
    \label{fig:depth}
    \vspace{-3mm}
\end{figure}

Furthermore, we present depth rendering results in Fig.~\ref{fig:depth}. Depth maps are computed using the default depth rendering technique~\cite{kerbl2024hierarchical} for 3DGS, without employing any depth regularization or additional supervision. For CityGaussian~\cite{liu2024citygaussian}, we utilize the official checkpoints. As shown in Fig.~\ref{fig:depth}, our method effectively captures fine-grained geometric details while avoiding floating artifacts. This demonstrates its strong ability to maintain multi-view consistency and preserve appearance information across a large set of images, ultimately enabling large-scale scene reconstruction with highly accurate geometric representation. These results further highlight the potential of our approach for future large-scale mesh reconstruction.

\end{document}